\pdfoutput=1
\documentclass[11pt,letterpaper,logo,onecolumn,copyright]{template_from_google}
\usepackage[authoryear,sort&compress,round]{natbib}
\usepackage{amsmath,amssymb,amsthm,bm,mathtools}
\usepackage{booktabs}
\usepackage{algorithm}
\usepackage{algorithmic}
\usepackage{multirow,pifont}
\usepackage{adjustbox}

\newcommand{\cmark}{\ding{51}}
\newcommand{\xmark}{\textcolor{gray!65}{\ding{55}}}
\definecolor{darkblue}{rgb}{0.0,0.0,0.6}
\definecolor{darkred}{rgb}{0.7,0.0,0.0}
\hypersetup{
  pdffitwindow=true,pdfstartview={FitH},pdfnewwindow=true,
  colorlinks=true,linktocpage=true,breaklinks=true,
  linkcolor=darkred,urlcolor=darkblue,citecolor=darkblue,
  pdftitle={Reinforcing Diffusion Models by Instantaneous Trajectory Distribution Matching},
  pdfauthor={Yihong Luo}
}
\usepackage[capitalize]{cleveref}
\makeatletter
\providecommand{\@LN}[2]{}
\providecommand{\@LN@col}[1]{}
\makeatother
\crefname{equation}{Eq.}{Eqs.}
\crefname{section}{Sec.}{Secs.}
\crefname{subsection}{Sec.}{Secs.}
\crefname{appendix}{Appendix}{Appendices}
\Crefname{appendix}{Appendix}{Appendices}
\crefname{subappendix}{Appendix}{Appendices}
\Crefname{subappendix}{Appendix}{Appendices}
\crefname{subsubappendix}{Appendix}{Appendices}
\Crefname{subsubappendix}{Appendix}{Appendices}
\crefname{table}{Tab.}{Tabs.}
\crefname{proposition}{Proposition}{Propositions}
\crefname{lemma}{Lemma}{Lemmas}
\crefname{remark}{Remark}{Remarks}
\crefname{algorithm}{Algorithm}{Algorithms}

\newcommand{\E}{\mathbb{E}}
\newcommand{\KL}{\operatorname{KL}}
\newcommand{\sg}{\operatorname{sg}}

\newcommand{\Normal}{\mathcal{N}}
\newcommand{\rvx}{\mathbf{x}}
\newcommand{\rvc}{\mathbf{c}}
\newcommand{\rvy}{\mathbf{y}}
\newcommand{\rvz}{\mathbf{z}}
\newcommand{\spara}[1]{\noindent\textbf{#1}}
\newtheorem{proposition}{Proposition}
\newtheorem{lemma}{Lemma}
\theoremstyle{remark}
\newtheorem{remark}{Remark}

\let\cite\citep

\title{Reinforcing Diffusion Models by Instantaneous Trajectory Distribution Matching}
\reportnumber{}
\renewcommand{\today}{}

\renewcommand{\titlefigure}{\noindent
\begin{minipage}{\linewidth}
    \centering
    \captionsetup{hypcap=false}
    \includegraphics[width=0.85\linewidth]{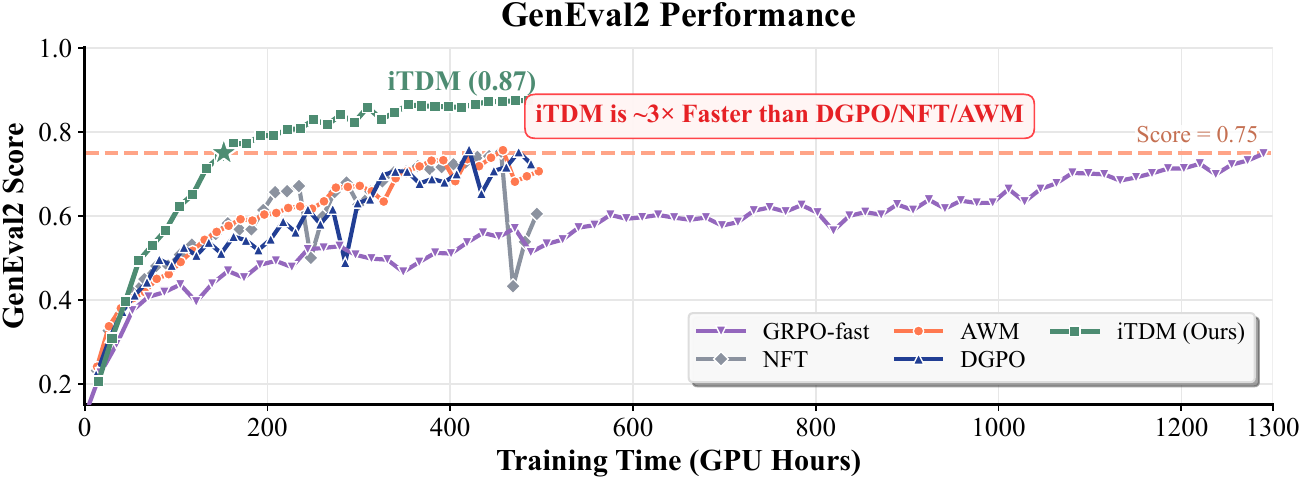}
    \captionof{figure}{\textbf{GenEval2 performance on SD3.5-M.} Our \textbf{iTDM} is nearly $3\times$ faster than DGPO, DiffusionNFT, and AWM, while continuously scaling to \textbf{0.87} without training collapse.}
    \label{fig:teaser}
\end{minipage}
\par\vspace{12pt}
}

\author{Yihong Luo}
\affil{Hong Kong University of Science and Technology}

\begin{abstract}
Forward diffusion reinforcement learning (RL)---enabling deterministic ODE sampling without trajectory backpropagation---has grown increasingly popular for efficiently aligning diffusion models. However, we reveal that existing methods essentially rely on signed reward-weighted regression (RWR), regressing the \emph{trainable policy} velocity onto rollouts from a \emph{behavior distribution}. Unbiased baseline subtraction implicitly assumes that the policy velocity equals the behavior marginal velocity---an idealized assumption violated under classifier-free guidance (CFG) rollouts, offline/off-policy updates, and imperfect online self-consistency. This \emph{behavior--policy mismatch} breaks baseline invariance and injects \emph{baseline-induced gradient drift}, destabilizing training. To resolve this issue, we propose \textbf{I}nstantaneous \textbf{T}rajectory \textbf{D}istribution \textbf{M}atching (\textbf{iTDM}), taking the infinitesimal limit of Trajectory Distribution Matching toward a time-local reward-guided target via the posterior-score identity to yield a sample-level gradient anchored at the behavior velocity without learning an additional reward-tilted target score. 
Tracking this velocity with an online auxiliary denoiser guarantees exact baseline invariance at the auxiliary optimum, effectively suppressing baseline drift across both online and offline RL. 
Extensive experiments across compositional generation, text rendering, and preference alignment show that iTDM achieves state-of-the-art in-domain and out-of-domain performance across group-wise online/offline RL and pairwise offline RL. 
Notably, on the challenging GenEval2 benchmark, iTDM converges nearly $3\times$ faster than competing methods while scaling smoothly to $0.87$ without collapse. Furthermore, iTDM scales effectively to the 5.6B FLUX.3 model, achieving better performance across both in-domain and unseen metrics. Code is available at \url{https://github.com/Luo-Yihong/iTDM}.
\end{abstract}

\begin{document}

\maketitle
\suppressfloats[t]

\section{Introduction}
\label{sec:intro}

\begin{figure}[!tbp]
    \centering
    \includegraphics[width=1.0\linewidth]{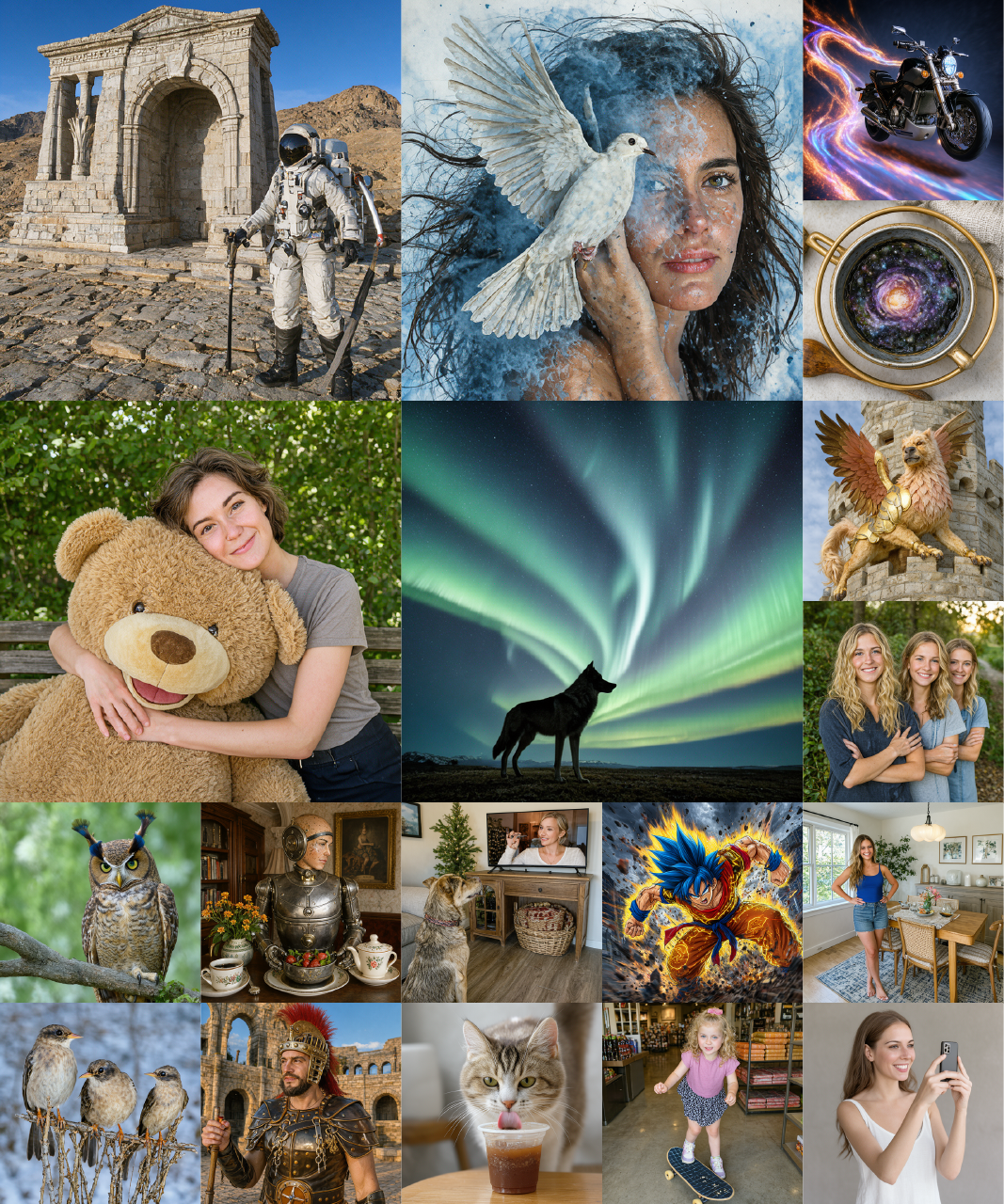}
    \caption{\textbf{Samples generated by iTDM on the 5.6B FLUX.3 Klein Base~\cite{flux3klein},} reinforced with HPSv3++~\cite{liu2026hpsv3pp} in the guidance-free regime ($\text{CFG}=1.0$).}
    \label{fig:flux3-itdm-showcase}
\end{figure}

\begin{figure}[!tbp]
    \centering
    \includegraphics[width=1.0\linewidth]{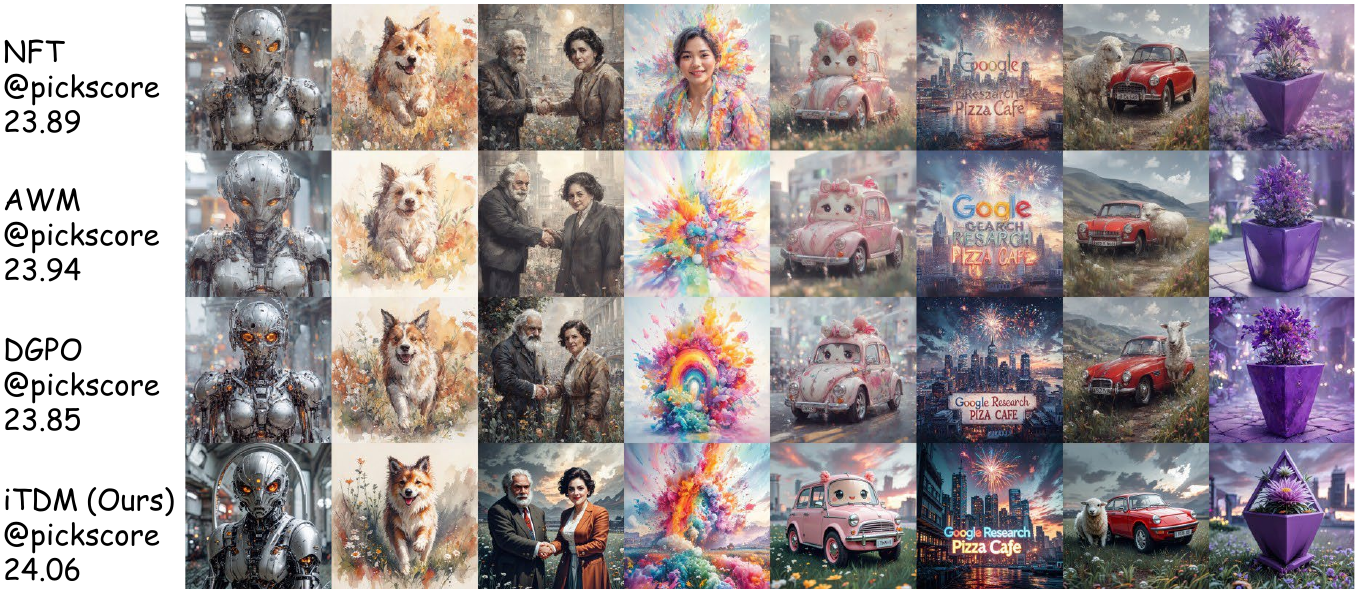}
    \caption{\textbf{Qualitative comparisons on human preference alignment (PickScore) with SD3.5-M.} All models are trained and evaluated under $\text{CFG}=1.0$. Compared with competing forward diffusion RL baselines, \textbf{iTDM} faithfully follows intricate prompt details while delivering sharper textures and superior aesthetic quality, \textbf{avoiding blurry artifacts}.}
    \label{fig:visual_compare}
\end{figure}

Reinforcement learning (RL) has become a cornerstone for post-training generative models, advancing large language models~\cite{ziegler2019fine,bai2022training,shao2024deepseekmath,deepseek_r1} and empowering diffusion and flow-matching models in compositional generation, text rendering, and preference alignment~\cite{black2023training,wallace2024diffusion,flowgrpo,dancegrpo}. Early online diffusion RL methods~\cite{flowgrpo,dancegrpo} apply trajectory-level policy gradients via stochastic differential equation (SDE) samplers, sacrificing the efficiency and fidelity of deterministic ordinary differential equation (ODE) solvers while incurring heavy trajectory-backpropagation overhead. Consequently, \emph{forward diffusion RL}~\cite{dgpo,rl_awm,rl_nft}---which collects clean rollouts via deterministic ODE solvers and optimizes policies along the forward noising path---has gained popularity for its training efficiency.

In GRPO-style post-training~\cite{shao2024deepseekmath}, subtracting a prompt-level baseline $b(\rvc)$ to form centered advantages ($A = R - b(\rvc)$) is essential for variance reduction and relative preference learning. However, we reveal that existing forward diffusion RL methods~\cite{dgpo,rl_nft,rl_awm} essentially rely on signed reward-weighted regression (RWR): the rollout sampler defines a \emph{behavior distribution} $q_0$ with marginal velocity $v_q(\rvx_t,t)$, whereas RWR evaluates its denoising residual at the \emph{trainable policy} velocity $v_\theta(\rvx_t,t)$. \emph{Unbiased baseline subtraction therefore implicitly rests on an idealized assumption: $v_\theta = v_q$}. Yet this assumption fails in practice under classifier-free guidance (CFG) rollouts, offline/off-policy updates, and imperfect online self-consistency, creating pervasive \emph{behavior--policy mismatch} ($v_\theta \neq v_q$) with two severe side effects:
(1) \emph{Baseline-induced gradient drift}: non-zero residual mean ($v_\theta - v_q \neq 0$) breaks baseline invariance and injects a spurious drift proportional to $b(\rvc)(v_\theta - v_q)$, causing unanchored RWR to collapse and existing methods to stagnate or diverge under CFG-guided and offline RL; and
(2) \emph{Coarse rollout-artifact fitting}: RWR's inherent pull toward $q_t$ drives $v_\theta$ to fit few-step ODE discretization artifacts, degrading textures and out-of-domain fidelity  (\cref{fig:visual_compare,fig:compare_cfg}).

We argue that these failures stem not from centered advantages, but from forcing distributional policy improvement into a signed regression tied to the fragile behavior-policy consistency assumption ($v_\theta = v_q$). Based on this insight, an ideal forward diffusion RL method should tolerate the effects of behavior--policy mismatch without drifting toward coarse few-step rollouts.

To this end, we propose \textbf{I}nstantaneous \textbf{T}rajectory \textbf{D}istribution \textbf{M}atching (\textbf{iTDM}), a principled distribution-matching framework for forward diffusion RL. Building on Trajectory Distribution Matching (TDM)~\cite{luo2025learning}, iTDM takes its \emph{infinitesimal limit} to match a reward-guided target distribution $\pi_t^R \propto q_t\exp(m_t)$ at each noise level. Via the Gaussian posterior-score identity, iTDM converts the intractable target score into an exact sample-level RL gradient anchored at the behavior velocity $v_q$, without learning an additional reward-tilted target score. Theoretically, iTDM ensures \emph{exact baseline invariance} even when $v_\theta \neq v_q$. Practically, we track $v_q$ online with an auxiliary denoiser $v_\phi$, guaranteeing exact baseline invariance at the auxiliary optimum and effectively suppressing baseline drift. Combined with pretrained reference regularization $v_{\mathrm{ref}}$, iTDM mitigates rollout-artifact fitting and yields a unique stationary policy independent of the relationship between $q_t$ and $v_\theta$, naturally unifying online and offline RL.

This design brings two key benefits:
(1) \textbf{Principled Drift and Artifact Resistance}, guaranteeing baseline invariance at the auxiliary optimum while decoupling regularization from coarse rollouts;
(2) \textbf{Stable Online and Offline Reinforcement Post-Training} across guided, unguided, group-wise, and pairwise setups.
Extensive experiments on SD3.5-M~\cite{sd3} across compositional generation, text rendering, and preference alignment validate iTDM's effectiveness: it converges nearly $3\times$ faster to \textbf{0.87} on GenEval2 without collapse (\cref{fig:teaser}), unlocks stable CFG-guided online RL with leading DrawBench out-of-domain quality, and overcomes offline collapse while outperforming Diffusion-DPO~\cite{wallace2024diffusion} in the pairwise offline setting ($G=2$). Beyond SD3.5-M, iTDM also scales to the 5.6B-parameter FLUX.3 Klein Base model~\cite{flux3klein} (\cref{fig:flux3-itdm-showcase,fig:flux3-itdm-compare}), where guidance-free reinforcement ($\text{CFG}=1.0$) with HPSv3++~\cite{liu2026hpsv3pp} surpasses the CFG-guided base model ($\text{CFG}=3.5$) across in-domain and unseen DrawBench metrics.

\section{Background}
\label{sec:prelim}

\spara{Diffusion Models and Flow Matching.}
Given a clean latent distribution $q_0$ (conditioned on a prompt $\rvc$, omitted when unambiguous) and independent Gaussian noise $\epsilon\sim\Normal(\mathbf{0},\mathbf{I})$, the linear forward interpolation path at noise level $t\in(0,1)$ and its sample-wise conditional velocity are
\begin{equation}
 \rvx_t=(1-t)\rvx_0+t\epsilon,\quad
 v(\rvx_t\mid\rvx_0):=\frac{d\rvx_t}{dt}=\epsilon-\rvx_0.
 \label{eq:path}
\end{equation}
Let $q_t$ denote the marginal distribution of $\rvx_t$. Regressing onto $v(\rvx_t\mid\rvx_0)$ yields the marginal velocity $v_q(\rvx,t)=\E[v(\rvx_t\mid\rvx_0)\mid\rvx_t=\rvx]$, linked to the marginal score via:
\begin{equation}
 s_{q_t}(\rvx)=-\frac{\rvx+(1-t)v_q(\rvx,t)}{t}.
 \label{eq:score-velocity}
\end{equation}

\spara{Policy Gradient and Baseline Invariance.}
In policy-based RL such as REINFORCE, PPO~\cite{schulman2017proximal}, and GRPO~\cite{shao2024deepseekmath}, a tractable stochastic policy $\pi_\theta(\rvx_0\mid\rvc)$ maximizes expected reward $\mathcal J(\theta)=\E_{\rvx_0\sim\pi_\theta(\cdot\mid\rvc)}[r(\rvx_0,\rvc)]$. 
A foundational property of policy gradients is \emph{baseline invariance}: subtracting any prompt-dependent baseline $b(\rvc)$ to form centered advantages $A(\rvx_0,\rvc)=r(\rvx_0,\rvc)-b(\rvc)$ leaves the population gradient strictly unchanged:
\begin{equation}
 \E_{\rvx_0\sim\pi_\theta(\cdot\mid\rvc)}\!\Big[\big(r(\rvx_0,\rvc)-b(\rvc)\big)\nabla_\theta\log\pi_\theta(\rvx_0\mid\rvc)\Big]
 =\nabla_\theta\mathcal J(\theta).
 \label{eq:prelim-pg-baseline}
\end{equation}
\textit{This exact baseline invariance legitimizes advantage centering in PPO and GRPO as a pure variance-reduction mechanism} without biasing the optimization direction.

\spara{Forward Diffusion RL.}
To bypass costly trajectory-level policy gradients, forward diffusion RL methods~\cite{dgpo,rl_nft,rl_awm} collect clean rollouts $\rvx_0\sim q_0$ from a \emph{behavior distribution} (e.g., an online rollout policy $p_{\theta^-}$ or frozen generator) via deterministic ODE solvers and optimize the \emph{trainable policy} velocity $v_\theta$ directly along the forward noising path. Up to gradient equivalence, their objectives unify as
\begin{equation}
 \mathcal L_{\mathrm{FwdRL}}(\theta)
 =\E_{\rvx_0\sim q_0,\,t,\epsilon}\Big[
   \sg(\omega)\big(r(\rvx_0,\rvc)-b(\rvc)\big)\|v(\rvx_t\mid\rvx_0)-v_\theta\|_2^2
 +\lambda\|v_\theta-v_{\mathrm{ref}}\|_2^2
   +\beta\|v_\theta-v_{\mathrm{old}}\|_2^2
 \Big],
 \label{eq:prelim-fwdrl}
\end{equation}
where $b(\rvc)$ is a prompt-level baseline, $v_{\mathrm{ref}}$ is the frozen pretrained model, and method-specific choices of $\omega=\omega(\rvx_0,\rvx_t)$, $v_{\mathrm{old}}$, and $(\lambda,\beta)$ are summarized in \cref{tab:prelim_fwdrl} (with $v_{\theta^-}$ denoting rollout-generating parameters).
At their core, existing methods rely on \emph{signed reward-weighted regression} (RWR), $(r-b)\|v(\rvx_t\mid\rvx_0)-v_\theta\|_2^2$, paired with different regularizers. While subtracting $b(\rvc)$ is indispensable for variance reduction, \textit{\textbf{we reveal that baseline subtraction in signed RWR is generally ill-posed under behavior--policy mismatch ($v_\theta\neq v_q$), violating baseline invariance and biasing the population gradient}}.

\begin{table}[!tbp]
\centering
\small
\caption{Unification of existing forward diffusion RL methods.}
\setlength{\tabcolsep}{4pt}
\begin{adjustbox}{max width=\linewidth}
\begin{tabular}{@{}lccc@{}}
\toprule
\textbf{Method} & \textbf{Weight $\omega$} & \textbf{Anchor $v_{\mathrm{old}}$} & \textbf{Default Regularization} \\
\midrule
DGPO~\cite{dgpo} & group-level sigmoid gate & --- & $\lambda\ge0,\ \beta=0$ \\
DiffusionNFT~\cite{rl_nft} & $1$ & $v_{\theta^-}$ & $\lambda\ge 0,\ \beta=(1 - r+b)$ \\
AWM~\cite{rl_awm} & Approx.\ importance ratio & $v_{\theta_{\mathrm{EMA}}}$ & $\lambda\ge 0,\ \beta=1$ \\
\bottomrule
\end{tabular}%
\end{adjustbox}
\label{tab:prelim_fwdrl}
\end{table}

\section{Method}
\label{sec:method}

Existing forward diffusion RL methods rely on signed RWR (\cref{eq:prelim-fwdrl}), which ties unbiased baseline subtraction to the fragile consistency assumption $v_\theta=v_q$ while pulling $v_\theta$ toward coarse rollout marginals. To tolerate behavior--policy mismatch without drifting toward rollout artifacts, we propose \textbf{I}nstantaneous \textbf{T}rajectory \textbf{D}istribution \textbf{M}atching (\textbf{iTDM}), which formulates forward diffusion RL by taking the \emph{infinitesimal limit} of Trajectory Distribution Matching (TDM)~\cite{luo2025learning}. Below, we derive the general instantaneous TDM gradient (\cref{sec:instantaneous}), construct a time-local reward-guided target that reduces to an exact sample-level update without learning an additional reward-tilted target score (\cref{sec:target}), establish exact baseline invariance alongside practical velocity estimation (\cref{sec:baseline_impl}; \cref{alg:itdm}), and theoretically compare iTDM with signed RWR (\cref{sec:awm}), with proofs and scope discussions in \cref{app:proofs_scope}.

\spara{Problem Setup.}
At each RL step, a rollout sampler (e.g., an online policy with or without CFG, or an offline generator) produces clean samples $\rvx_0$ under prompt $\rvc$, inducing a \emph{behavior distribution} $q_0$ (held fixed with respect to $\theta$ within each update) with bounded nonnegative reward $R=r(\rvx_0,\rvc)$. Along the forward path in \cref{eq:path}, $v_q(\rvx,t)=\E[v(\rvx_t\mid\rvx_0)\mid\rvx_t=\rvx]$ is the behavior marginal velocity and $m_t(\rvx)=\E[R\mid\rvx_t=\rvx]$ is the conditional expected reward. Crucially, the \emph{trainable policy} velocity $v_\theta(\rvx,t)$ generally differs from $v_q(\rvx,t)$ ($v_\theta\neq v_q$)---a \emph{behavior--policy mismatch} arising under CFG-guided rollouts, offline or off-policy updates, and imperfect online self-consistency.

\subsection{Instantaneous Trajectory Distribution Matching}
\label{sec:instantaneous}

To perform forward diffusion RL without backpropagating through multi-step solvers, we consider a local reverse transition from $t$ to $s=t-\delta$ ($\delta>0$). Following TDM~\cite{luo2025learning}, for fixed $(t,\delta)$ we define the one-step transport map $T_\theta$ and its pushforward marginal $q_s^\theta$:
\begin{equation}
 T_\theta(\rvx)=\rvx-\delta v_\theta(\rvx,t),\qquad
 q_s^\theta=(T_\theta)_\#q_t.
 \label{eq:transport}
\end{equation}
For any smooth target marginal $p_s^\star$ held fixed during the update, minimizing the $1/\delta$-scaled reverse KL objective $\mathcal K_\delta(\theta):=\frac{1}{\delta}\KL(q_s^\theta\|p_s^\star)$ with $\rvx_s=T_\theta(\rvx_t)$ yields the pathwise transport gradient:
\begin{equation}
 \nabla_\theta\mathcal K_\delta(\theta)
 =\frac{1}{\delta}\E_{q_t}\!\left[
  \left(\frac{\partial\rvx_s}{\partial\theta}\right)^\top
  \big(s_{q_s^\theta}(\rvx_s)-s_{p_s^\star}(\rvx_s)\big)
 \right].
 \label{eq:transport-gradient}
\end{equation}
Since $\frac{\partial\rvx_s}{\partial\theta}=-\delta J_\theta$ at fixed $\rvx_t$, where $J_\theta:=\frac{\partial v_\theta(\rvx_t,t)}{\partial\theta}$ denotes the policy Jacobian, taking the infinitesimal limit $\delta\downarrow 0$ yields the \textbf{\emph{instantaneous TDM}} gradient:
\begin{equation}
 \boxed{
 g_t^\star
 :=\lim_{\delta\downarrow0}\nabla_\theta\mathcal K_\delta(\theta)
 =\E_{q_t}\!\big[
 J_\theta^\top
 (s_{p_t^\star}-s_{q_t})\big].}
 \label{eq:instantaneous}
\end{equation}
Here arguments $(\rvx_t,t)$ are suppressed when unambiguous. With \cref{eq:instantaneous} adapting TDM to standard multi-step diffusion and flow-matching models, the key remaining challenge for RL is specifying a tractable reward-guided target score $s_{p_t^\star}$.

\subsection{Reward-Guided Instantaneous Matching}
\label{sec:target}

We now instantiate $p_t^\star$ for reinforcement learning. At each noise level $t$, we set $p_t^\star$ to the time-local reward-guided target $\pi_t^R$ with score $s_t^R$:
\begin{equation}
 \begin{gathered}
 \pi_t^R(\rvx)=\frac{q_t(\rvx)e^{m_t(\rvx)}}{Z_t},\quad
 Z_t=\E_{q_t}[e^{m_t(\rvx_t)}],\\
 \boxed{s_t^R(\rvx):=s_{\pi_t^R}(\rvx)
                 =s_{q_t}(\rvx)+\nabla_{\rvx} m_t(\rvx).}
 \end{gathered}
 \label{eq:target}
\end{equation}
With bounded rewards, $\pi_t^R$ uniquely solves the time-local KL-regularized variational problem $\max_p\{\E_p[m_t]-\KL(p\|q_t)\}$, where $\nabla_{\rvx} m_t$ serves as the reward-driven score correction.

\spara{Sample-Level RL Gradient.}
Directly evaluating $s_t^R(\rvx)$ is intractable, as $m_t(\rvx)=\E[R\mid\rvx_t=\rvx]$ is an unknown conditional expectation over clean latents that cannot be differentiated directly. Fortunately, $\nabla_{\rvx} m_t(\rvx)$ admits an exact conditional expectation form:

\begin{lemma}[Posterior-score identity for reward correction]
\label{lem:posterior-score}
Let $q_t(\rvx\mid\rvx_0)=\Normal(\rvx;(1-t)\rvx_0,t^2\mathbf{I})$ denote the Gaussian transition kernel of \cref{eq:path}. The reward-driven score correction $\nabla_{\rvx} m_t(\rvx)$ satisfies
\begin{align}
 \nabla_{\rvx} m_t(\rvx)
 &=\E\!\left[R\big(\nabla_{\rvx}\log q_t(\rvx\mid\rvx_0)-s_{q_t}(\rvx)\big)
             \mid\rvx_t=\rvx\right] \notag\\
 &=\frac{1-t}{t}\E\!\left[R\big(v_q(\rvx_t,t)-v(\rvx_t\mid\rvx_0)\big)\mid\rvx_t=\rvx\right].
 \label{eq:posterior-score}
\end{align}
\end{lemma}

Yet \cref{lem:posterior-score} alone is not immediately practical: substituting $s_t^R(\rvx_t)$ into \cref{eq:instantaneous} still involves a conditional expectation given $\rvx_t$, which would ordinarily require learning an additional reward-tilted target score network onto \cref{eq:posterior-score}. Crucially, in the instantaneous limit, the iTDM gradient reduces directly to a tractable sample-level update without learning an additional reward-tilted target score (proof in \cref{app:proofs}):

\begin{proposition}[Sample-level gradient of reward-guided instantaneous matching]
\label{prop:instantaneous}
Under the regularity assumptions in \cref{app:proofs}, the time-weighted instantaneous gradient for the reward-guided target $\pi_t^R$ in \cref{eq:target} satisfies
\begin{equation}
 \boxed{\begin{aligned}
 g_t^R&:=\frac{2t}{1-t}\lim_{\delta\downarrow0}
  \nabla_\theta\mathcal K_\delta(\theta)\\
 &=2\E\!\left[R\,
 J_\theta^\top
 \big(v_q-v(\rvx_t\mid\rvx_0)\big)\right].
 \end{aligned}}
 \label{eq:normalized-gradient}
\end{equation}
\end{proposition}

\begin{remark}[Sample-level target estimation does not extend to few-step TDM]
\label{rem:tower-measurability}
Because \cref{prop:instantaneous} avoids learning an additional reward-tilted target score via a simple sample-level update, one might ask whether few-step TDM~\cite{luo2025learning,luo2026tdmr1,luo2026vatdm} admits the same trick. Crucially, this reduction holds \emph{only} in the instantaneous limit: with $\rvx_t$ held fixed, $J_\theta$ depends solely on $\rvx_t$, allowing the outer expectation over $q_t$ to absorb the conditional expectation given $\rvx_t$ via the tower property. In few-step generation, however, the generator Jacobian at a re-noised state remains coupled with the clean sample prediction, invalidating single-sample substitution (see \cref{app:proofs}).
\end{remark}

\spara{Regularized iTDM Objective and Optimal Solution.}
For reinforcement post-training, we combine $g_t^R$ with reference-velocity regularization weighted by $\beta>0$ to form the iTDM-RL gradient:
\begin{equation}
 g_t^{\mathrm{iTDM}}
 =g_t^R+\beta\nabla_\theta\E_{q_t}\!\left[\big\|v_\theta(\rvx_t,t)-v_{\mathrm{ref}}(\rvx_t,t)\big\|_2^2\right].
 \label{eq:reg-gradient-vA}
\end{equation}

\begin{lemma}[Optimal policy and time-local policy improvement]
\label{lem:optimal-reg-a}
For any behavior marginal $q_t$ with full support, frozen reference marginal $p_{\mathrm{ref},t}$ with score $s_t^{\mathrm{ref}}$, and $\beta>0$, the unique pointwise stationary velocity $v^\star$ of \cref{eq:reg-gradient-vA} and its score $s_t^\star$ are
\begin{equation}
 \begin{aligned}
 v^\star(\rvx,t)&=v_{\mathrm{ref}}(\rvx,t)-\frac{t}{\beta(1-t)}\nabla_{\rvx} m_t(\rvx),\\
 s_t^\star(\rvx)&=s_t^{\mathrm{ref}}(\rvx)+\frac{1}{\beta}\nabla_{\rvx} m_t(\rvx).
 \end{aligned}
 \label{eq:optimal-solution-a}
\end{equation}
Moreover, $s_t^\star=\nabla_{\rvx}\log\pi_t^+$ for $\pi_t^+(\rvx)=p_{\mathrm{ref},t}(\rvx)e^{m_t(\rvx)/\beta}/Z_t^\beta$ with $Z_t^\beta=\E_{p_{\mathrm{ref},t}}[e^{m_t(\rvx_t)/\beta}]$. Defining $\mathcal J_t(p):=\E_p[m_t]-\beta\KL(p\|p_{\mathrm{ref},t})$ with $m_t$ held fixed, if $\KL(q_t\|p_{\mathrm{ref},t})<\infty$, then $\pi_t^+$ achieves the exact improvement
\begin{equation}
 \mathcal J_t(\pi_t^+)-\mathcal J_t(q_t)=\beta\KL(q_t\|\pi_t^+)\ge 0,
 \label{eq:local-policy-improvement}
\end{equation}
which is strict whenever $q_t\neq\pi_t^+$.
\end{lemma}
\cref{lem:optimal-reg-a} reveals two structural advantages of regularized iTDM: first, the stationary target $\pi_t^+$ strictly improves the time-local regularized reward over $q_t$ whenever $q_t\neq\pi_t^+$; second, $s_t^\star$ is anchored at $s_t^{\mathrm{ref}}$ and its form is independent of the relationship between the behavior distribution $q_t$ and trainable policy $v_\theta$. Consequently, \textit{\textbf{regularized iTDM naturally unifies online and offline RL}}.

\begin{figure}[!tbp]
    \centering
    \includegraphics[width=1.0\linewidth]{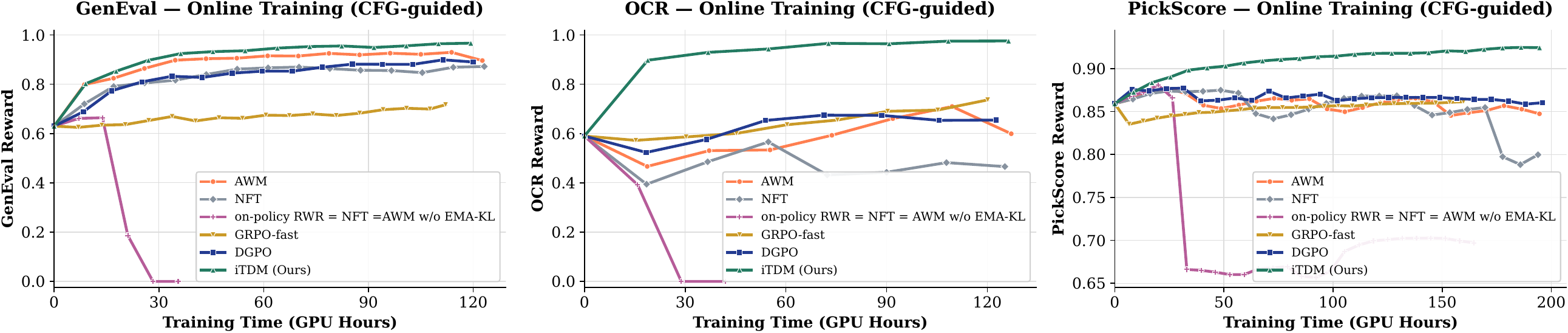}
    \caption{\textbf{Online reinforcement post-training efficiency with CFG-Guided policy.}  Note that \textbf{on-policy RWR} (equal to on-policy AWM without EMA anchors and on-policy NFT) rapidly stagnates or degrades under baseline-induced drift, whereas \textbf{iTDM} converges substantially faster and reaches higher asymptotic rewards without late-stage collapse.}
    \label{fig:online_cfg45_hours}
\end{figure}

\subsection{Baseline Invariance and Practical Learning}
\label{sec:baseline_impl}

Following GRPO~\cite{shao2024deepseekmath,flowgrpo}, for each prompt $\rvc$ we sample a group of $G$ clean rollouts $\{\rvx_{0,i}\}_{i=1}^G\sim q_0$ and subtract a prompt-dependent baseline $b(\rvc)$ to form centered advantages $A=R-b(\rvc)$ for variance reduction and relative preference learning. A key question is whether subtracting $b(\rvc)$ alters the population iTDM gradient.

\begin{lemma}[Exact baseline invariance]
\label{lem:baseline-invariance}
For any prompt-dependent baseline $b(\rvc)$, subtracting $b(\rvc)$ leaves the population iTDM gradient strictly invariant:
\begin{equation}
 \boxed{g_t^A
 :=2\E\!\left[A\,J_\theta^\top\big(v_q-v(\rvx_t\mid\rvx_0)\big)\right]
 =g_t^R.}
 \label{eq:baseline}
\end{equation}
\end{lemma}
Crucially, \textit{\textbf{exact baseline invariance holds even under behavior--policy mismatch ($v_\theta\neq v_q$)}}. This yields two practical benefits: 1) \textbf{Stable Offline and Off-Policy Training}: it natively supports unbiased advantage centering under static datasets, EMA-lagged or multi-epoch replay rollouts, and CFG-guided sampling; and 2) \textbf{Robustness to Imperfect Online Self-Consistency}: even in purely online rollouts generated by $v_\theta$, the forward marginal velocity $v_q$ generally differs from $v_\theta$ before full convergence to the optimal velocity field~\cite{song2020score,liu2022flow,karras2022elucidating}.

\spara{Learning Auxiliary Velocity $v_q$.}
Since the ground-truth behavior velocity $v_q$ is not analytically available, we introduce an online auxiliary denoiser $v_\phi$ to estimate $v_q$, following score distillation~\cite{yin2023one,luo2025learning,luo2026vatdm}. Noisy states $\rvx_t$ are constructed by re-noising terminal rollouts $\rvx_0\sim q_0$ via \cref{eq:path} without storing intermediate solver trajectories, and $v_\phi$ (parameterized as a lightweight LoRA adapter) minimizes the standard denoising loss:
\begin{equation}
 \mathcal L_{\mathrm{aux}}(\phi)
 =\E_{t,\rvx_0,\epsilon}\!\left[\big\|v_\phi(\rvx_t,t)-v(\rvx_t\mid\rvx_0)\big\|_2^2\right],
 \label{eq:dsm}
\end{equation}
whose unique minimizer is $v_\phi=v_q$. In practice, we substitute the stop-gradient prediction $v_\phi$ for $v_q$, yielding the practical gradient $\widehat g_t^A=2\E[A\,J_\theta^\top(v_\phi-v(\rvx_t\mid\rvx_0))]$, and center rewards with the group-mean baseline $b(\rvc)=\frac{1}{G}\sum_{i=1}^G R_i$, combining $\widehat g_t^A$ with the reference regularization in \cref{eq:reg-gradient-vA} to optimize $v_\theta$ (\cref{alg:itdm}).

\begin{algorithm}[!tbp]
\caption{iTDM for Forward Diffusion RL}
\label{alg:itdm}
\small
\begin{algorithmic}[1]
\REQUIRE Policy $v_\theta$, auxiliary $v_\phi$, reference $v_{\mathrm{ref}}$, reward $r$, prompts $\mathcal{D}_c$, group size $G$, weight $\beta$.
\FOR{$n \leftarrow 1$ {\bfseries to} $N$}
\STATE Sample $\rvc \sim \mathcal{D}_c$ and $\{\rvx_{0,i}\}_{i=1}^G \sim q_0(\cdot\mid\rvc)$; set $R_i \gets r(\rvx_{0,i}, \rvc)$, $A_i \gets \operatorname{clip}(\widehat{\operatorname{Adv}}_i, -5, 5)/5$
\STATE Sample $t \sim \mu$, $\epsilon_i \sim \Normal(\mathbf{0}, \mathbf{I})$, $\rvx_{t,i} \gets (1-t)\rvx_{0,i} + t\epsilon_i$
\STATE Update $\phi$ by minimizing $\mathcal{L}_{\mathrm{aux}}(\phi)$ in \cref{eq:dsm}
\STATE Update $\theta$ along $-g_t^{\mathrm{iTDM}}$ (\cref{eq:reg-gradient-vA}) with $g_t^R \gets \widehat g_t^A$ using $\sg(v_\phi)$
\ENDFOR
\end{algorithmic}
\end{algorithm}

\subsection{Theoretical Analysis}
\label{sec:awm}

To isolate the core mechanism shared by existing forward diffusion RL methods (\cref{eq:prelim-fwdrl}), we consider a \emph{fully online} setting ($\omega=1$) and analyze the core RWR objective at noise level $t$ with weight $w\in\{R,A\}$:
\begin{equation}
 \mathcal L_{\mathrm{RWR}}^{w}(\theta)
 =\E\!\left[w\,\big\|v(\rvx_t\mid\rvx_0)-v_\theta(\rvx_t,t)\big\|_2^2\right].
 \label{eq:rwr}
\end{equation}
Whereas nonnegative weighting ($w=R\ge 0$) admits a well-defined pointwise minimizer $\E[R\,v(\rvx_t\mid\rvx_0)\mid\rvx_t=\rvx]/m_t(\rvx)$, centering rewards into signed advantages $w=A=R-b(\rvc)$ renders $\mathcal L_{\mathrm{RWR}}^A$ unbounded below in general. 
To examine whether baseline subtraction remains valid in RWR and clarify what signed RWR optimizes, we characterize its population gradient:

\begin{proposition}[Baseline non-invariance and gradient decomposition of RWR]
\label{prop:awm-drift}
For any prompt-dependent baseline $b(\rvc)$ and centered advantage $A=R-b(\rvc)$, the population gradient of RWR in \cref{eq:rwr} satisfies:

\noindent 1) \textbf{Violation of baseline invariance:} Subtracting $b(\rvc)$ shifts the population gradient of RWR by
\begin{equation}
 \boxed{
  \nabla_\theta\mathcal L_{\mathrm{RWR}}^{A}
  -\nabla_\theta\mathcal L_{\mathrm{RWR}}^{R}
  =-\Delta_t^b,}
 \label{eq:awm-baseline}
\end{equation}
where $\Delta_t^b:=2b(\rvc)\,\E_{q_t}\big[J_\theta^\top(v_\theta-v_q)\big]$ is the baseline-induced drift.

\noindent 2) \textbf{Gradient decomposition:} The population gradient of $\mathcal L_{\mathrm{RWR}}^A$ decomposes into
\begin{equation}
 \nabla_\theta\mathcal L_{\mathrm{RWR}}^{A}
 =\underbrace{g_t^A}_{\text{iTDM gradient}}
 +\underbrace{2\E_{q_t}\!\big[
       m_t\,J_\theta^\top(v_\theta-v_q)\big]}_{\Delta_t^q:\ \text{drift toward } q_t}
   -\Delta_t^b.
 \label{eq:awm-decomposition}
\end{equation}
\end{proposition}

\spara{Side Effects of Signed RWR.}
\cref{prop:awm-drift} isolates two extraneous drift terms in signed RWR beyond $g_t^A$:
1) \textbf{\emph{Baseline-induced gradient drift ($-\Delta_t^b$)}}: whenever $v_\theta\neq v_q$, subtracting $b(\rvc)>0$ pushes $v_\theta$ \emph{away} from $v_q$ during gradient descent, causing unanchored on-policy RWR to collapse (\cref{fig:online_cfg45_hours,fig:online_cfg1_hours}) and anchored variants to stagnate or diverge under CFG-guided and offline rollouts (\cref{tab:drawbench_online,fig:offline_compare_hours}); and
2) \textbf{\emph{Coarse rollout-artifact fitting ($\Delta_t^q$)}}: since $m_t\ge 0$, $\Delta_t^q$ pulls $v_\theta$ toward $v_q$, forcing online policies to fit coarse few-step ODE artifacts~\cite{flowgrpo,dgpo} and degrading visual and out-of-domain quality even at $\text{CFG}=1.0$ (\cref{fig:visual_compare,tab:drawbench_online}).

\spara{Advantages of iTDM.}
At the auxiliary optimum ($v_\phi=v_q$), anchoring the residual at $v_q$ while regularizing toward $v_{\mathrm{ref}}$ instead of $q_t$ eliminates both $-\Delta_t^b$ and $\Delta_t^q$ (\cref{app:proofs_scope}). This simultaneously insulates $v_\theta$ from few-step rollout artifacts for superior visual fidelity (\cref{fig:visual_compare,tab:drawbench_online}) and preserves unbiased advantage centering (\cref{lem:baseline-invariance}) for stable optimization across guided, unguided, and offline regimes without restrictive proximity anchors.

\begin{remark}
\label{rem:aux-approx}
With the learned denoiser $v_\phi$, practical baseline invariance holds up to $v_\phi - v_q$ (\cref{app:scope}). Yet unlike signed RWR---which passively assumes $v_\theta = v_q$ even as mismatch and RL updates violate it---iTDM decouples velocity estimation from policy improvement and \emph{actively minimizes} $\|v_\phi - v_q\|_2^2$ (\cref{eq:dsm}). 
Approximating intractable generator scores with an online auxiliary network has been extensively validated across score distillation~\cite{wang2023prolificdreamer,yin2023one,zhou2024score,dmd2,luo2025learning,luo2025adding,luo2026vatdm}. 
Despite maintaining an auxiliary adapter, iTDM achieves markedly greater stability, less distributional drift, and faster wall-clock convergence than RWR-based methods (\cref{fig:teaser,fig:online_cfg45_hours,fig:online_cfg1_hours}).
\end{remark}

\section{Experiments}
\label{sec:exp}

\subsection{Experimental Setup}
\label{sec:exp_setup}

\spara{Evaluation Tasks.}
We post-train SD3.5-M~\cite{sd3} using LoRA adapters~\cite{hu2022lora} for both $v_\theta$ and $v_\phi$ across three benchmarks~\cite{flowgrpo,dgpo}: 1) \emph{Compositional Image Generation} on GenEval~\cite{ghosh2023geneval} and the harder GenEval2~\cite{kamath2025geneval2} (\cref{fig:teaser}); 2) \emph{Visual Text Rendering} (OCR)~\cite{chen2023textdiffuser,gong2025seedream} for character-level accuracy; and 3) \emph{Human Preference Alignment} with PickScore~\cite{kirstain2023pick}. To further assess scalability across model architectures, parameter scales, and reward functions, we also apply iTDM to the 5.6B-parameter FLUX.3 Klein Base model~\cite{flux3klein} using the wide-spectrum human preference reward model HPSv3++~\cite{liu2026hpsv3pp}.

\spara{Training Setups and Baselines.}
We evaluate \textbf{online RL} under both \emph{CFG-guided} ($\text{CFG}=4.5$) and \emph{conditional} ($\text{CFG}=1.0$) policies, alongside frozen-rollout \textbf{offline RL} in \emph{group-wise} ($G>2$) and \emph{pairwise} ($G=2$) settings. We compare against forward diffusion RL baselines---DGPO~\cite{dgpo}, DiffusionNFT (NFT)~\cite{rl_nft}, and AWM~\cite{rl_awm} (plus unanchored \textbf{on-policy RWR})---Diffusion-DPO (DPO)~\cite{wallace2024diffusion} at $G=2$, and trajectory-based GRPO-fast~\cite{flowgrpo} for efficiency reference.

\spara{Out-of-Domain Evaluation Metrics.}
To guard against reward hacking, we evaluate seven unseen quality and preference metrics on DrawBench~\cite{saharia2022photorealistic} (see \cref{app:exp_detail} for details).

\subsection{Online Reinforcement Post-Training}
\label{sec:exp_online}

\begin{figure}[!tbp]
    \centering
    \includegraphics[width=1.0\linewidth]{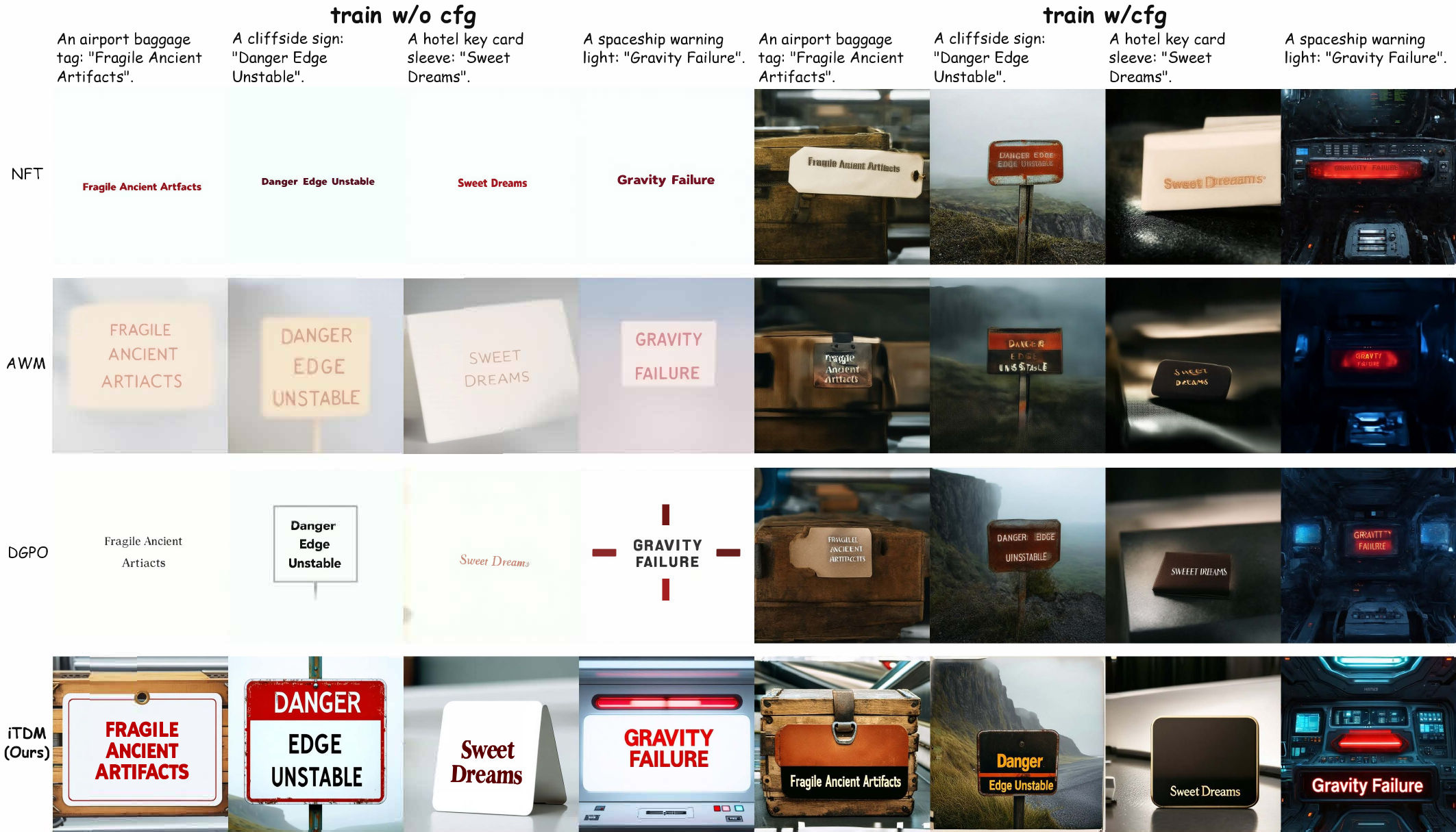}
    \caption{\textbf{Visual text rendering under conditional vs.\ CFG-guided online RL.} Without CFG (left), baselines overfit text accuracy by collapsing onto plain, context-free backgrounds; with CFG (right), they preserve scene context but fail to render legible text due to velocity-mismatch drift. \textbf{iTDM} excels in both regimes, uniting rich scene fidelity with sharp, accurate text under CFG-guided RL.}
    \label{fig:compare_cfg}
\end{figure}

\spara{Necessity of CFG-Guided Policy RL.}
Classifier-free guidance is vital for exploration and visual fidelity: removing CFG sharply degrades base SD3.5-M ($0.63\to 0.24$ GenEval, $0.59\to 0.12$ OCR in \cref{tab:drawbench_online}), and unguided RL ($\text{CFG}=1.0$) easily sacrifices scene context despite high task rewards (left panel of \cref{fig:compare_cfg}). Yet reinforcing a CFG-guided policy stalls RWR baselines due to persistent behavior--policy mismatch (\cref{prop:awm-drift}), yielding blurry or illegible text ($0.70$--$0.76$ OCR; right panel of \cref{fig:compare_cfg}). Tracking $v_q$ via $v_\phi$ resolves this dilemma and unlocks stable CFG-guided RL.

\begin{table}[!tbp]
    \centering
    \renewcommand{\arraystretch}{1.04}
    \caption{\textbf{Online reinforcement post-training with guided ($\text{CFG}=4.5$, \cmark) and unguided ($\text{CFG}=1.0$, \xmark) policy.} All methods use the same strength of regularization toward the frozen reference. Best results in each setting are highlighted in \textbf{bold}.}
    \begin{adjustbox}{max width=\linewidth}
        \begin{tabular}{clcccccccccc}
            \toprule
            \multirow{2}{*}{\textbf{CFG}} & \multirow{2}{*}{\textbf{Model}}
            & \multicolumn{3}{c}{\textbf{Task Metric}}
            & \multicolumn{2}{c}{\textbf{Image Quality}}
            & \multicolumn{5}{c}{\textbf{Alignment \& Preference}} \\
            \cmidrule(lr){3-5} \cmidrule(lr){6-7} \cmidrule(l){8-12}
            & & \textbf{GenEval} & \textbf{OCR} & \textbf{PickScore}
            & \textbf{Aesthetic} & \textbf{DeQA}
            & \textbf{CLIPScore} & \textbf{HPSv2.1}
            & \textbf{ImgRwd} & \textbf{PickScore} & \textbf{UniRwd} \\

            \midrule
            \cmark & SD3.5-M$^\dagger$ & 0.63 & 0.59 & 21.72 & 5.36 & --- & 28.50 & 27.90 & 0.85 & 22.34 & 3.03 \\
            \xmark & SD3.5-M$^\dagger$ & 0.24 & 0.12 & 19.82 & 5.13 & --- & 23.70 & 20.40 & -0.58 & 20.51 & 2.02 \\

            \midrule
            \multicolumn{12}{l}{\textbf{\textit{Compositional Image Generation:}}} \\
            \midrule
            \multirow{5}{*}{\cmark}
            & AWM & 0.93 & --- & --- & 5.12 & 3.43 & 28.96 & 24.54 & 0.91 & 22.14 & 3.07 \\
            & NFT & 0.89 & --- & --- & 4.92 & 2.97 & 28.74 & 22.62 & 0.64 & 21.89 & 2.97 \\
            & DGPO & 0.90 & --- & --- & 5.09 & 3.17 & 28.98 & 23.87 & 0.75 & 22.05 & 2.98 \\
            & \textbf{iTDM (Ours)} & \textbf{0.97} & --- & --- & \textbf{5.26} & \textbf{4.12} & \textbf{29.00} & \textbf{28.19} & \textbf{0.98} & \textbf{22.49} & \textbf{3.13} \\
            \cmidrule(lr){1-12}
            \multirow{4}{*}{\xmark}
            & AWM & 0.92 & --- & --- & 4.71 & 3.94 & 27.50 & 21.41 & 0.15 & 21.35 & 2.62 \\
            & NFT & 0.96 & --- & --- & 4.88 & 4.05 & 27.34 & 22.90 & 0.30 & 21.51 & 2.82 \\
            & DGPO & 0.97 & --- & --- & \textbf{5.15} & 4.06 & 27.64 & 24.03 & \textbf{0.71} & \textbf{22.04} & 2.96 \\
            & \textbf{iTDM (Ours)} & \textbf{0.98} & --- & --- & 5.08 & \textbf{4.23} & \textbf{27.69} & \textbf{24.73} & 0.65 & 21.74 & \textbf{3.04} \\

            \midrule
            \multicolumn{12}{l}{\textbf{\textit{Visual Text Rendering:}}} \\
            \midrule
            \multirow{5}{*}{\cmark}
            & AWM  & --- & 0.75 & --- & \textbf{5.28} & 3.62 & 28.67 & 24.77 & 0.73 & 22.05 & 2.93 \\
            & NFT  & --- & 0.70 & --- & 5.19 & 3.67 & 28.75 & 25.43 & 0.68 & 22.10 & 2.91 \\
            & DGPO & --- & 0.76 & --- & 4.95 & 2.56 & 26.82 & 19.73 & -0.45 & 20.76 & 2.18 \\
            & \textbf{iTDM (Ours)} & --- & \textbf{0.96} & --- & 5.23 & \textbf{3.97} & \textbf{28.77} & \textbf{27.98} & \textbf{0.80} & \textbf{22.40} & \textbf{3.09} \\
            \cmidrule(lr){1-12}
            \multirow{4}{*}{\xmark}
            & AWM & --- & 0.85 & --- & 4.16 & 3.25 & 21.02 & 13.70 & -1.69 & 19.69 & 1.52 \\
            & NFT & --- & \textbf{0.97} & --- & 3.52 & 3.52 & 22.13 & 14.86 & -1.49 & 19.86 & 1.50 \\
            & DGPO & --- & \textbf{0.97} & --- & 4.19 & 2.13 & 21.36 & 14.16 & -1.70 & 19.69 & 1.53 \\
            & \textbf{iTDM (Ours)} & --- & \textbf{0.97} & --- & \textbf{4.48} & \textbf{3.88} & \textbf{22.80} & \textbf{18.25} & \textbf{-1.15} & \textbf{20.14} & \textbf{1.82} \\

            \midrule
            \multicolumn{12}{l}{\textbf{\textit{Human Preference Alignment:}}} \\
            \midrule
            \multirow{5}{*}{\cmark}
            & AWM & --- & --- & 22.80 & 6.02 & 3.99 & 27.73 & 29.78 & 1.18 & 23.23 & 3.27 \\
            & NFT & --- & --- & 22.75 & 5.96 & 3.86 & 27.50 & 28.72 & 1.14 & 23.14 & 3.25 \\
            & DGPO & --- & --- & 22.37 & 5.65 & 3.98 & 28.12 & 28.87 & 1.10 & 22.89 & 3.17 \\
            & \textbf{iTDM (Ours)} & --- & --- & \textbf{24.08} & \textbf{6.11} & \textbf{4.12} & \textbf{28.35} & \textbf{32.73} & \textbf{1.46} & \textbf{24.26} & \textbf{3.52} \\
            \cmidrule(lr){1-12}
            \multirow{4}{*}{\xmark}
            & AWM & --- & --- & 23.94 & \textbf{6.45} & 3.97 & 26.59 & 29.99 & 1.29 & 23.95 & 3.34 \\
            & NFT & --- & --- & 23.89 & 6.40 & 3.81 & 26.36 & 29.05 & 1.24 & 23.86 & 3.32 \\
            & DGPO & --- & --- & 23.85 & 6.18 & 3.60 & \textbf{27.32} & 29.21 & 1.23 & 23.54 & 3.35 \\
            & \textbf{iTDM (Ours)} & --- & --- & \textbf{24.06} & 6.22 & \textbf{4.26} & 26.71 & \textbf{31.85} & \textbf{1.33} & \textbf{24.00} & \textbf{3.36} \\
            \bottomrule
        \end{tabular}%
    \end{adjustbox}
    \label{tab:drawbench_online}
\end{table}

\spara{Quantitative Results.}
In \cref{tab:drawbench_online}, under the \textbf{CFG-guided policy} ($\text{CFG}=4.5$, \cmark), iTDM leads all benchmarks (\textbf{0.97} GenEval, \textbf{0.96} OCR, \textbf{24.08} PickScore) while staying on par with base SD3.5-M on DrawBench and sweeping all seven unseen metrics on PickScore ($6.11$ Aesthetic, $32.73$ HPSv2.1, $1.46$ ImgRwd). In contrast, RWR baselines trade out-of-domain quality for task reward: on GenEval, NFT and DGPO fall to $22.62$ and $23.87$ HPSv2.1 (vs.\ $27.90$ for the base model and \textbf{28.19} for iTDM), and on OCR, DGPO drops to $19.73$ HPSv2.1 and $-0.45$ ImgRwd despite longer training while still trailing iTDM by $0.20$ in OCR accuracy. On PickScore, iTDM surpasses the strongest baseline AWM by $1.28$ in-domain ($24.08$ vs.\ $22.80$) and by $2.95$ HPSv2.1 out-of-domain ($32.73$ vs.\ $29.78$), indicating that its reward gains reflect genuine preference improvement rather than reward hacking. Consistent with \cref{prop:awm-drift}, this gap reflects the baseline-induced drift $-\Delta_t^b$ under CFG-guided rollouts.

Under the \textbf{conditional policy} ($\text{CFG}=1.0$, \xmark), coarse rollout-artifact fitting ($\Delta_t^q$ in \cref{prop:awm-drift}) causes RWR baselines on OCR to lose perceptual quality despite $0.97$ accuracy ($3.52$ Aesthetic for NFT; $2.13$ DeQA for DGPO), whereas iTDM matches or tops all three tasks (\textbf{0.98}, \textbf{0.97}, \textbf{24.06}) with markedly higher DrawBench fidelity, e.g., \textbf{18.25} vs.\ at most $14.86$ HPSv2.1 on OCR and the best DeQA on every task.

\begin{figure}[!tbp]
    \centering
    \includegraphics[width=1.0\linewidth]{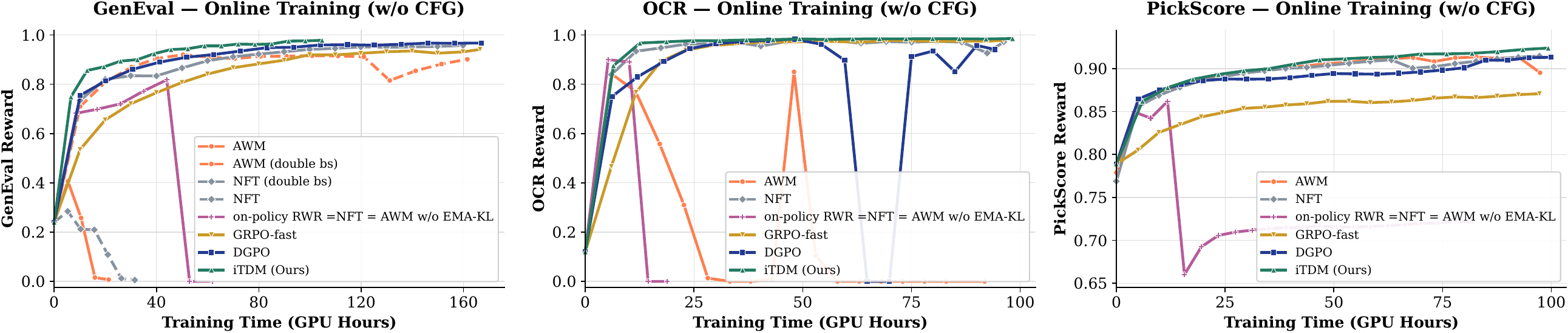}
    \caption{\textbf{Online reinforcement post-training efficiency without guidance ($\text{CFG}=1.0$).} Note that \textbf{on-policy RWR} is equal to on-policy AWM without EMA anchors and on-policy NFT.
    We double the batch size for AWM and NFT to stabilize training.}
    \label{fig:online_cfg1_hours}
\end{figure}

\spara{Training Cost and Scaling.}
The lightweight auxiliary LoRA $v_\phi$ adds negligible overhead, enabling iTDM to converge substantially faster in GPU hours than GRPO-fast and RWR baselines (\cref{fig:online_cfg45_hours,fig:online_cfg1_hours}). On the harder \textbf{GenEval2} benchmark (\cref{fig:teaser}), \textit{\textbf{iTDM reaches $0.75$ nearly $3\times$ faster than DGPO, NFT, and AWM}}, scaling smoothly to \textbf{0.87} without collapse.

\spara{Qualitative Comparison.}
Qualitative comparisons on PickScore (\cref{fig:visual_compare}) and OCR (\cref{fig:compare_cfg}) show that RWR baselines suffer from blurry textures, collapsed backgrounds, or distorted text, whereas iTDM delivers crisp details, rich scene context, and accurate text rendering.

\begin{table}[!tbp]
    \centering
    \renewcommand{\arraystretch}{1.06}
    \caption{\textbf{Scaling online reinforcement post-training to the 5.6B FLUX.3 Klein Base~\cite{flux3klein}.} Reinforced with HPSv3++~\cite{liu2026hpsv3pp} under guidance-free sampling ($\text{CFG}=1.0$). Best results are highlighted in \textbf{bold}.}
    \begin{adjustbox}{max width=\linewidth}
        \begin{tabular}{lcccccccc}
            \toprule
            \multirow{2}{*}{\textbf{Model}}
            & \textbf{Task Metric}
            & \multicolumn{2}{c}{\textbf{Image Quality}}
            & \multicolumn{5}{c}{\textbf{Alignment \& Preference}} \\
            \cmidrule(lr){2-2} \cmidrule(lr){3-4} \cmidrule(l){5-9}
            & \textbf{HPSv3++}
            & \textbf{Aesthetic} & \textbf{DeQA}
            & \textbf{CLIPScore} & \textbf{HPSv2.1}
            & \textbf{ImgRwd} & \textbf{PickScore} & \textbf{UniRwd} \\
            \midrule
            FLUX.3 Klein Base ($\text{CFG}=1.0$) & 1.66 & 5.06 & 3.86 & 23.13 & 20.84 & -0.65 & 20.46 & 2.13 \\
            FLUX.3 Klein Base ($\text{CFG}=3.5$) & 8.52 & 5.31 & 4.27 & \textbf{28.28} & 27.14 & 0.73 & \textbf{22.33} & 3.26 \\
            \midrule
            \textbf{iTDM (Ours, $\text{CFG}=1.0$)} & \textbf{10.92} & \textbf{5.44} & \textbf{4.46} & 26.18 & \textbf{27.70} & \textbf{0.83} & 22.17 & \textbf{3.38} \\
            \bottomrule
        \end{tabular}%
    \end{adjustbox}
    \label{tab:flux3-itdm-drawbench}
\end{table}

\begin{figure}[!tbp]
    \centering
    \includegraphics[width=1.0\linewidth]{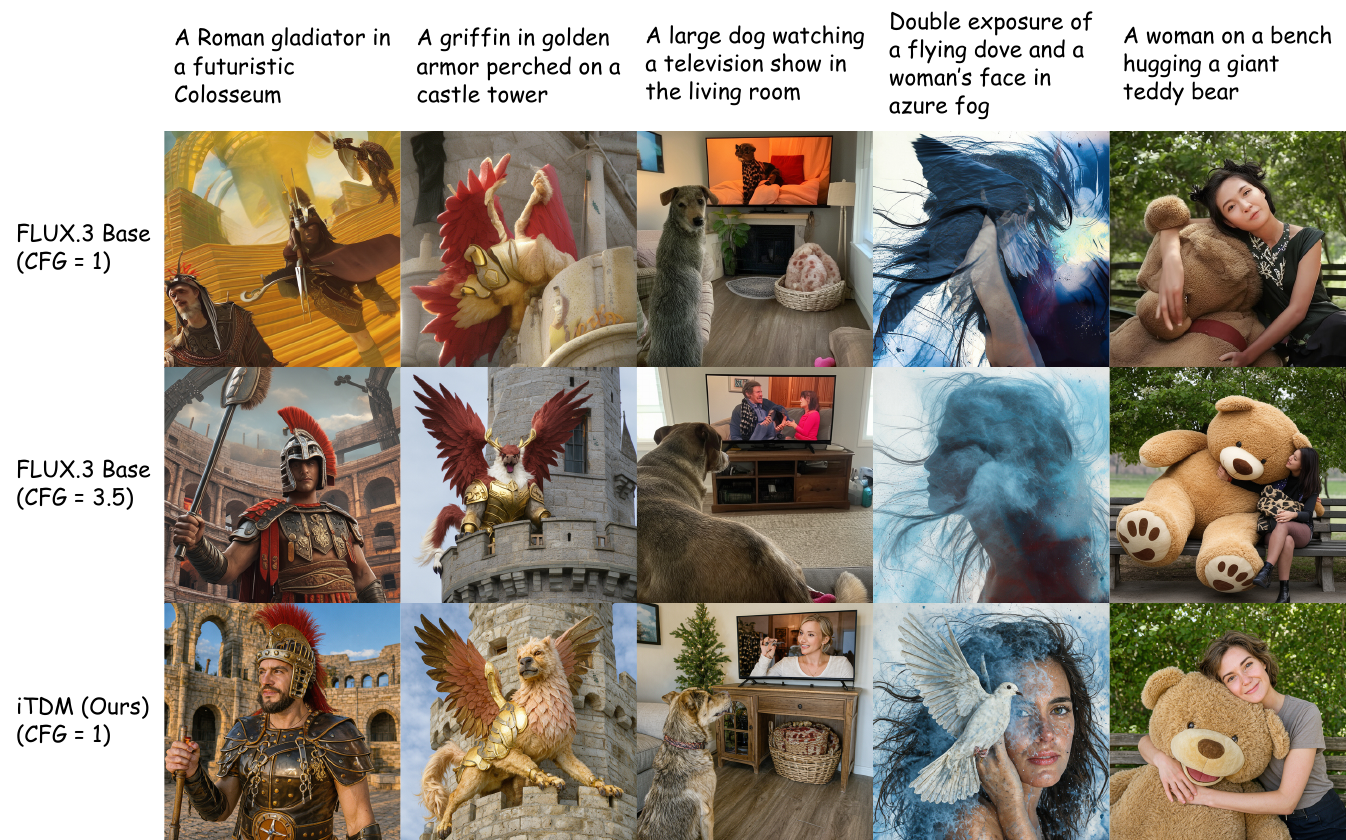}
    \caption{\textbf{Qualitative comparison between the 5.6B FLUX.3 Klein Base and iTDM.} Reinforced under $\text{CFG}=1.0$ with HPSv3++, \textbf{iTDM} delivers sharper visual fidelity and more faithful adherence to complex compositional prompts than the base model at both $\text{CFG}=1.0$ and $\text{CFG}=3.5$.}
    \label{fig:flux3-itdm-compare}
\end{figure}

\spara{Scaling to 5.6B FLUX.3 Klein Base.}
To further evaluate scalability across model architectures, parameter scales, and reward functions, we apply iTDM to the larger 5.6B-parameter FLUX.3 Klein Base~\cite{flux3klein} using HPSv3++~\cite{liu2026hpsv3pp}---a wide-spectrum reward model calibrated across model capability levels and RL iterations---as the training signal in the guidance-free regime ($\text{CFG}=1.0$). As shown in \cref{tab:flux3-itdm-drawbench}, compared with the unguided base model and guided base model, guidance-free iTDM ($\text{CFG}=1.0$) achieves gains in most out-of-domain metrics, confirming genuine improvements in perceptual fidelity and semantic alignment rather than reward hacking. Qualitative results in \cref{fig:flux3-itdm-showcase,fig:flux3-itdm-compare} corroborate these gains, demonstrating sharper visual details, richer lighting and textures, and more faithful adherence to complex compositional prompts.

\subsection{Offline RL and Diagnostic Analysis}
\label{sec:exp_offline}

\begin{figure}[!tbp]
    \centering
    \includegraphics[width=1.0\linewidth]{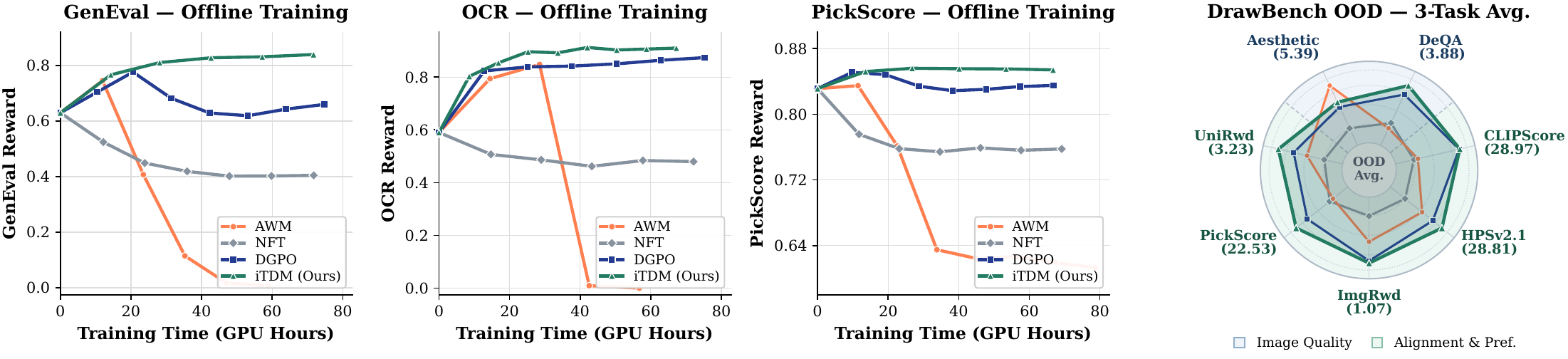}
    \caption{\textbf{Group-wise offline RL with frozen rollouts ($\text{CFG}=4.5$).} \emph{Left:} Task reward vs.\ GPU hours. \emph{Right:} Seven unseen DrawBench metrics averaged over the three tasks (parentheses denote \textbf{iTDM}). \textbf{iTDM} improves monotonically without collapse and leads out-of-domain generalization.}
    \label{fig:offline_compare_hours}
\end{figure}

\spara{Group-Wise Frozen-Rollout RL.}
With $q_0$ frozen to pretrained SD3.5-M ($\text{CFG}=4.5$), using the initial velocity $v_{\theta_0}$ as a static proxy in offline AWM and NFT leaves persistent mismatch ($v_{\theta_0}\neq v_q$; \cref{fig:offline_compare_hours,tab:drawbench_offline}). Consequently, NFT collapses below the base model ($0.52$ vs.\ $0.63$ GenEval; $0.53$ vs.\ $0.59$ OCR) and AWM diverges after peaking at $0.75$ GenEval. Supported by its decoupled stationary policy (\cref{lem:optimal-reg-a}) and baseline invariance at the auxiliary optimum (\cref{lem:baseline-invariance}), iTDM improves monotonically to \textbf{0.87} GenEval ($+9\%$ over DGPO), \textbf{0.92} OCR, and \textbf{22.26} PickScore while leading DrawBench generalization.

\begin{table}[!tbp]
    \centering
    \renewcommand{\arraystretch}{1.06}
    \caption{\textbf{Offline training across pairwise ($G=2$) and group-wise settings.} Rollouts are generated by the frozen pretrained SD3.5-M ($\text{CFG}=4.5$); all methods use the same strength of regularization toward the frozen reference. Best results in each setting are in \textbf{bold}.}
    \begin{adjustbox}{max width=\linewidth}
        \begin{tabular}{lcccccccccc}
            \toprule
            \multirow{2}{*}{\textbf{Model}} & \multicolumn{3}{c}{\textbf{Task Metric}} & \multicolumn{2}{c}{\textbf{Image Quality}} & \multicolumn{5}{c}{\textbf{Alignment \& Preference}} \\
            \cmidrule(lr){2-4} \cmidrule(lr){5-6} \cmidrule(l){7-11}
            & \textbf{GenEval} & \textbf{OCR} & \textbf{PickScore} & \textbf{Aesthetic} & \textbf{DeQA} & \textbf{CLIPScore} & \textbf{HPSv2.1} & \textbf{ImgRwd} & \textbf{PickScore} & \textbf{UniRwd} \\

            \midrule
            SD3.5-M$^\dagger$ & 0.63 & 0.59 & --- & 5.36 & --- & 28.50 & 27.90 & 0.85 & 22.34 & 3.03 \\
            \midrule
            \multicolumn{11}{l}{\textbf{\textit{Compositional Image Generation:}}} \\
            \midrule
            DPO ($G=2$) & 0.69 & --- & --- & \textbf{5.41} & 4.03 & 28.72 & \textbf{28.55} & 0.97 & 22.35 & \textbf{3.20} \\
            \textbf{iTDM (Ours, $G=2$)} & \textbf{0.77} & --- & --- & 5.38 & \textbf{4.06} & \textbf{29.11} & 28.26 & \textbf{1.06} & \textbf{22.43} & 3.18 \\
            \cmidrule(lr){1-11}
            AWM & 0.75 & --- & --- & \textbf{5.36} & 3.72 & 29.10 & 27.66 & 0.93 & 21.49 & 3.14 \\
            NFT & 0.52 & --- & --- & 5.33 & 3.76 & 28.30 & 27.13 & 0.72 & 22.00 & 3.00 \\
            DGPO & 0.78 & --- & --- & 5.33 & \textbf{3.99} & \textbf{29.11} & \textbf{28.01} & 1.01 & 22.28 & 3.21 \\
            \textbf{iTDM (Ours)} & \textbf{0.87} & --- & --- & 5.32 & 3.82 & 28.93 & 27.98 & \textbf{1.03} & \textbf{22.38} & \textbf{3.23} \\

            \midrule
            \multicolumn{11}{l}{\textbf{\textit{Visual Text Rendering:}}} \\
            \midrule
            DPO ($G=2$) & --- & 0.78 & --- & 5.23 & 3.90 & \textbf{29.44} & \textbf{27.99} & 0.91 & 22.13 & 3.08 \\
            \textbf{iTDM (Ours, $G=2$)} & --- & \textbf{0.87} & --- & \textbf{5.29} & \textbf{3.95} & 29.26 & 27.57 & \textbf{1.01} & \textbf{22.23} & \textbf{3.13} \\
            \cmidrule(lr){1-11}
            AWM & --- & 0.85 & --- & \textbf{5.34} & 3.48 & 28.92 & 27.64 & 0.90 & 21.82 & 3.01 \\
            NFT & --- & 0.53 & --- & 5.21 & 3.52 & 28.33 & 25.97 & 0.62 & 21.74 & 2.95 \\
            DGPO & --- & 0.87 & --- & 5.18 & 3.33 & \textbf{29.41} & 26.37 & \textbf{0.99} & 21.90 & 2.97 \\
            \textbf{iTDM (Ours)} & --- & \textbf{0.92} & --- & 5.22 & \textbf{3.68} & 29.27 & \textbf{27.69} & 0.95 & \textbf{22.27} & \textbf{3.12} \\

            \midrule
            \multicolumn{11}{l}{\textbf{\textit{Human Preference Alignment:}}} \\
            \midrule
            DPO ($G=2$) & --- & --- & 21.87 & 5.46 & 4.06 & 28.44 & 28.77 & 0.98 & 22.51 & 3.16 \\
            \textbf{iTDM (Ours, $G=2$)} & --- & --- & \textbf{22.11} & \textbf{5.56} & \textbf{4.12} & \textbf{28.67} & \textbf{30.25} & \textbf{1.17} & \textbf{22.77} & \textbf{3.25} \\
            \cmidrule(lr){1-11}
            AWM & --- & --- & 21.71 & \textbf{5.67} & 3.73 & 27.02 & 27.55 & 0.83 & 22.20 & 3.09 \\
            NFT & --- & --- & 20.17 & 5.29 & 3.74 & 28.23 & 26.69 & 0.65 & 21.96 & 3.02 \\
            DGPO & --- & --- & 22.13 & 5.59 & \textbf{4.18} & 28.36 & 30.39 & 1.16 & 22.78 & 3.27 \\
            \textbf{iTDM (Ours)} & --- & --- & \textbf{22.26} & 5.62 & 4.15 & \textbf{28.71} & \textbf{30.75} & \textbf{1.24} & \textbf{22.94} & \textbf{3.34} \\
            \bottomrule
        \end{tabular}%
    \end{adjustbox}
    \label{tab:drawbench_offline}
\end{table}

\begin{figure}[!tbp]
    \centering
    \includegraphics[width=1.0\linewidth]{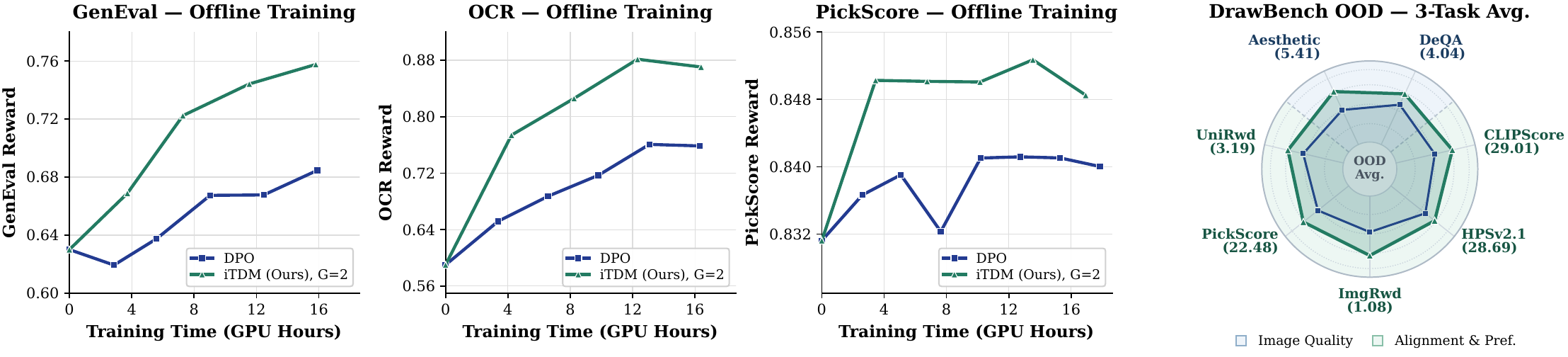}
    \caption{\textbf{Pairwise offline RL ($G=2$, $\text{CFG}=4.5$) vs.\ Diffusion-DPO.} \emph{Left:} Task reward vs.\ GPU hours. \emph{Right:} Three-task average DrawBench radar (parentheses denote \textbf{iTDM}). Even at $G=2$, \textbf{iTDM} converges faster and surpasses DPO across all tasks and unseen metrics.}
    \label{fig:offline_dpo_hours}
\end{figure}

\spara{Pairwise RL ($G=2$) and Group Scaling.}
Unlike pairwise-only DPO~\cite{wallace2024diffusion}, iTDM supports arbitrary group sizes. Under identical pairwise rollouts ($G=2$, \cref{fig:offline_dpo_hours}), iTDM converges faster and outperforms DPO on all tasks (\textbf{0.77} vs.\ $0.69$ GenEval, \textbf{0.87} vs.\ $0.78$ OCR, \textbf{22.11} vs.\ $21.87$ PickScore) and all seven unseen metrics. Scaling to larger groups (\cref{fig:offline_compare_hours,tab:drawbench_offline}) further boosts performance ($0.77\to 0.87$ GenEval, $0.87\to 0.92$ OCR).

\spara{Diagnostic Controls.}
\cref{fig:online_cfg45_hours,fig:online_cfg1_hours} verify the crucial control: removing proximity anchors collapses \textbf{on-policy RWR}, confirming that NFT and AWM rely on heuristic anchors to damp baseline drift, whereas iTDM suppresses it via $v_\phi$ with better rewards and training stability.

\section{Related Work}
\label{sec:related}

Aligning diffusion and flow-matching models with human preferences or downstream rewards generally follows three paradigms: supervised fine-tuning on curated image--prompt pairs~\cite{dai2023emu,podell2023sdxl}, explicit reward maximization via multi-step output backpropagation~\cite{prabhudesai2023aligning,clark2023directly,lee2023aligning,luo2025reward,luo2025adding}, control/score matching~\cite{psenka2023learning,domingo2025adjoint,lee2026rsm}, or policy gradients~\cite{fan2024reinforcement,black2023training,ye2024schedule,ren2024diffusion}, and implicit preference learning such as Diffusion-DPO~\cite{wallace2024diffusion}, Diffusion-KTO~\cite{li2024diffusionkto}, and group-enumerated variants~\cite{chen2025selfimprovementdiffusionmodelsgroup}. Recently, Flow-GRPO~\cite{flowgrpo} and DanceGRPO~\cite{dancegrpo} adapted GRPO~\cite{shao2024deepseekmath} to visual generation, inspiring refinements in hybrid sampling, branching, scheduling, and video alignment~\cite{mixgrpo,li2026branchgrpo,he2026tempflow,wang2025grpoguard,wang2025prefgrpo,liang2026leapalign,wang2026tagrpo,li2025taros}. However, these trajectory-based methods remain tied to stochastic transition likelihoods and costly multi-step trajectory training.

To enable fast deterministic ODE rollouts without trajectory backpropagation, recent forward diffusion RL methods optimize policies directly on independently re-noised endpoints. Beyond group-level preference optimization in DGPO~\cite{dgpo}, DiffusionNFT~\cite{rl_nft}, AWM~\cite{rl_awm}, EPG/PEPG~\cite{choi2026rethinking}, and RAM~\cite{bergmeister2026ram} (alongside~\cite{ye2025datareg,xu2026designing}) formulate velocity-space updates, which RVM~\cite{choi2026scaling} unifies at the gradient level as reward-weighted velocity regression with anchor regularization. Yet, because these RWR-style losses evaluate the residual at $v_\theta$, under CFG rollouts, offline/off-policy updates, or imperfect online self-consistency ($v_\theta \neq v_q$), subtracting a reward baseline violates baseline invariance and induces population gradient drift (\cref{sec:awm}). By taking the step-normalized infinitesimal limit of Trajectory Distribution Matching~\cite{luo2025learning}, \textbf{iTDM} anchors the sample-level residual at the behavior velocity via a lightweight auxiliary denoiser, guaranteeing exact baseline invariance and drift-free optimization at the auxiliary optimum to enable stable online and offline RL.

\section{Conclusion}
\label{sec:conclusion}

In this work, we revealed that existing forward diffusion RL methods essentially rely on signed RWR, whose implicit $v_\theta = v_q$ assumption breaks down under behavior--policy mismatch to induce baseline gradient drift while pulling policies toward coarse few-step rollout artifacts. To resolve both issues, we introduced \textbf{Instantaneous Trajectory Distribution Matching} (\textbf{iTDM}), which takes the infinitesimal limit of TDM toward a time-local reward-guided target via the posterior-score identity, yielding an exact sample-level RL gradient anchored at the behavior velocity without learning an additional reward-tilted target score. Paired with a lightweight online auxiliary LoRA denoiser and reference regularization, iTDM guarantees exact baseline invariance at the auxiliary optimum, suppresses drift and artifact fitting, and unifies online and offline RL. Extensive experiments across SD3.5-M and the 5.6B FLUX.3 model demonstrate up to $3\times$ faster convergence alongside state-of-the-art in-domain and out-of-domain performance across diverse tasks and post-training regimes.

\bibliography{main}

\clearpage
\appendix

\section{Detailed Proofs and Scope Discussion}
\label{app:proofs_scope}

\subsection{Regularity Assumptions and Proofs}
\label{app:proofs}

\spara{Regularity Assumptions.}
We consider interior times $t\in(0,1)$ with $\E\|\rvx_0\|_2^2<\infty$ and bounded rewards, holding $q_0$, $b(\rvc)$, and $\phi$ fixed with respect to $\theta$ during each update. We assume $v_\theta(\rvx,t)$ is continuously differentiable in $(\rvx,\theta)$ such that $T_\theta$ in \cref{eq:transport} is a diffeomorphism for sufficiently small $\delta>0$, that $q_s^\theta$ and $p_s^\star$ are positive and smooth with scores converging to $s_{q_t}$ and $s_{p_t^\star}$ as $\delta\downarrow 0$, and that vanishing boundary terms and integrable domination allow interchanging differentiation and limits with expectations.

\begin{proof}[Proof of \cref{lem:posterior-score}, \cref{prop:instantaneous}, and \cref{rem:tower-measurability}]
The $1/\delta$-weighted reverse-KL transport derivative is given by \cref{eq:transport-gradient}. Substituting $\frac{\partial\rvx_s}{\partial\theta}=-\delta J_\theta$ and taking the dominated limit as $\delta\downarrow 0$ proves \cref{eq:instantaneous}. Next, for the Gaussian kernel $q_t(\rvx_t\mid\rvx_0)=\Normal(\rvx_t;(1-t)\rvx_0,t^2\mathbf{I})$ in \cref{eq:path}, its conditional score satisfies
\begin{equation}
 \nabla_{\rvx}\log q_t(\rvx_t\mid\rvx_0)=-\frac{\epsilon}{t}
 =-\frac{\rvx_t+(1-t)v(\rvx_t\mid\rvx_0)}{t}.
\end{equation}
Differentiating the numerator and denominator of the Bayes representation of $m_t(\rvx)=\E[R\mid\rvx_t=\rvx]$ yields the first line of \cref{eq:posterior-score}. Subtracting the marginal score--velocity relation in \cref{sec:prelim} from $\nabla_{\rvx}\log q_t(\rvx_t\mid\rvx_0)$ gives the second line of \cref{eq:posterior-score}. 

To establish the single-sample reduction in \cref{eq:normalized-gradient} (\cref{rem:tower-measurability}), note that with $\rvx_t$ held fixed in the instantaneous limit $\delta\downarrow 0$, the normalized derivative $\delta^{-1}\partial\rvx_s/\partial\theta=-J_\theta$ is measurable with respect to $\rvx_t$. By the tower property of conditional expectation, for any integrable $f(\rvx_t,\rvx_0)$,
\begin{equation}
 \E_{\rvx_t}\!\left[J_\theta^\top\E[f(\rvx_t,\rvx_0)\mid\rvx_t]\right]
 =\E_{\rvx_t,\rvx_0}\!\left[J_\theta^\top f(\rvx_t,\rvx_0)\right].
 \label{eq:tower-measurability}
\end{equation}
Setting $f(\rvx_t,\rvx_0)=R(v_q(\rvx_t,t)-v(\rvx_t\mid\rvx_0))$ and multiplying by $\frac{2t}{1-t}$ yields \cref{eq:normalized-gradient}. In contrast, for a finite-step generator $\rvy=G_\theta(\rvz)+\sigma\epsilon$ ($\sigma>0$) with conditional score $s_\theta(\rvy\mid\rvz)=-(\rvy-G_\theta(\rvz))/\sigma^2$, the Jacobian $\partial\rvy/\partial\theta$ retains dependence on $\rvz$ given $\rvy$, so $\E[(\frac{\partial\rvy}{\partial\theta})^\top\E[s_\theta(\rvy\mid\rvz)\mid\rvy]]\neq\E[(\frac{\partial\rvy}{\partial\theta})^\top s_\theta(\rvy\mid\rvz)]$ in general.
\end{proof}

\begin{proof}[Proof of \cref{lem:optimal-reg-a}]
Using \cref{eq:posterior-score,eq:normalized-gradient}, conditioning on $\rvx_t=\rvx$ inside $g_t^{\mathrm{iTDM}}$ (\cref{eq:reg-gradient-vA}) yields the pointwise integrand $2J_\theta^\top[\frac{t}{1-t}\nabla_{\rvx} m_t(\rvx)+\beta(v_\theta(\rvx,t)-v_{\mathrm{ref}}(\rvx,t))]$. For any $q_t$ with full support, setting the bracketed vector to zero yields $v^\star(\rvx,t)=v_{\mathrm{ref}}(\rvx,t)-\frac{t}{\beta(1-t)}\nabla_{\rvx} m_t(\rvx)$, and applying the score--velocity relation $s_t(\rvx)=-\frac{\rvx+(1-t)v(\rvx,t)}{t}$ (\cref{sec:prelim}) gives $s_t^\star(\rvx)=s_t^{\mathrm{ref}}(\rvx)+\frac{1}{\beta}\nabla_{\rvx} m_t(\rvx)$ in \cref{eq:optimal-solution-a}. Hence $s_t^\star(\rvx)=\nabla_{\rvx}\log\pi_t^+(\rvx)$. Moreover, any density $p$ with $\KL(p\|p_{\mathrm{ref},t})<\infty$ satisfies $\mathcal J_t(p)=\beta\log Z_t^\beta-\beta\KL(p\|\pi_t^+)$; subtracting this identity at $p=q_t$ from $p=\pi_t^+$ yields \cref{eq:local-policy-improvement}, with strictness from the positivity of KL divergence.
\end{proof}

\begin{proof}[Proof of \cref{lem:baseline-invariance,prop:awm-drift}]
Conditioning on $(\rvx_t,\rvc)$ fixes the prompt baseline $b(\rvc)$ and the Jacobian $J_\theta$, while by definition $\E[v_q(\rvx_t,t)-v(\rvx_t\mid\rvx_0)\mid\rvx_t,\rvc]=0$. The baseline contribution to $g_t^A$ therefore vanishes identically, proving $g_t^A=g_t^R$ in \cref{lem:baseline-invariance} (\cref{eq:baseline}). For RWR, conditioning its residual gives $\E[v_\theta-v(\rvx_t\mid\rvx_0)\mid\rvx_t,\rvc]=v_\theta-v_q$; multiplying by $-2b(\rvc)$ and $J_\theta^\top$ proves the baseline shift $-\Delta_t^b$ in \cref{prop:awm-drift} (\cref{eq:awm-baseline}). Finally, adding and subtracting $v_q$ inside $v_\theta-v(\rvx_t\mid\rvx_0)$, conditioning $(R-b(\rvc))(v_\theta-v_q)$ on $(\rvx_t,\rvc)$ to obtain $(m_t-b(\rvc))(v_\theta-v_q)$, and separating $m_t$ from $-b(\rvc)$ prove the decomposition in \cref{eq:awm-decomposition}.
\end{proof}

\subsection{Approximation and Implementation Scope}
\label{app:scope}

\spara{Auxiliary Approximation Error and Baseline Dependence.}
In practice, $v_\phi$ approximates $v_q$ with residual error $e_\phi=v_\phi-v_q$, so $\widehat g_t^A=2\E[A\,J_\theta^\top(v_\phi-v(\rvx_t\mid\rvx_0))]$ satisfies $\widehat g_t^A-g_t^A=2\E[A\,J_\theta^\top e_\phi]$. By the Cauchy--Schwarz inequality, the gradient approximation error is bounded by
\begin{equation}
 \|\widehat g_t^A-g_t^A\|
 \leq 2
 \sqrt{\E\!\left[A^2\|J_\theta\|_{\mathrm{op}}^2\right]}
 \sqrt{\E\!\left[\|e_\phi\|_2^2\right]},
\end{equation}
and baseline invariance holds up to $e_\phi$ (with $\widehat g_t^R$ denoting $\widehat g_t^A$ at $A=R$):
\begin{equation}
 \widehat g_t^A-\widehat g_t^R
 =-2b(\rvc)\,\E\!\left[J_\theta^\top e_\phi\right].
\end{equation}
Minimizing $\mathcal L_{\mathrm{aux}}(\phi)$ in \cref{eq:dsm} directly drives $\E[\|e_\phi\|_2^2]$ toward zero, simultaneously suppressing both the gradient bias and residual baseline shift.

\spara{Group Baselines, Normalization, and Clipping.}
Whereas \cref{lem:baseline-invariance} considers a fixed prompt baseline $b(\rvc)$, drawing $G$ independent rollouts $\{\rvx_{0,i}\}_{i=1}^G$ for prompt $\rvc$ defines the per-sample zero-mean vector
\begin{equation}
 h_i=\left(\frac{\partial v_\theta(\rvx_{t,i},t)}{\partial\theta}\right)^\top
 \big(v_q(\rvx_{t,i},t)-v(\rvx_{t,i}\mid\rvx_{0,i})\big),
\end{equation}
with $\E[h_i\mid\rvc]=0$. By conditional independence across $j\neq i$, centering with the empirical group mean $\overline{R}=\frac{1}{G}\sum_{j=1}^G R_j$ yields
\begin{equation}
 \E[(R_i-\overline R)h_i\mid\rvc]
 =\left(1-\frac{1}{G}\right)\E[R_i h_i\mid\rvc],
\end{equation}
scaling the unbiased gradient by the constant factor $(1-1/G)$ (absorbable into the learning rate, or removed via the leave-one-out mean $\overline{R}_{-i}=\frac{1}{G-1}\sum_{j\neq i}R_j$). Moreover, group-level standard-deviation normalization and clipping (\cref{alg:itdm}) act as bounded reward shaping: for any induced joint law of $(\rvx_0,A)$ under independent forward noising, the posterior-score identity applies identically toward the shaped target $\pi_t^A\propto q_t e^{\E[A\mid\rvx_t,\rvc]}$.

\section{Experimental Details}
\label{app:exp_detail}

\spara{Task Descriptions and Reward Signals.}
Following Flow-GRPO~\cite{flowgrpo} and DGPO~\cite{dgpo}, we evaluate post-training on SD3.5-M~\cite{sd3} across three tasks:
\begin{itemize}
    \item \emph{Compositional Image Generation}: GenEval~\cite{ghosh2023geneval} assesses six compositional skills (single/two-object generation, counting, color attribution, spatial relations, and attribute binding), while GenEval2~\cite{kamath2025geneval2} (\cref{fig:teaser}) tests scalability under denser multi-attribute constraints.
    \item \emph{Visual Text Rendering (OCR)}: Following TextDiffuser~\cite{chen2023textdiffuser} and Seedream~\cite{gong2025seedream}, prompts use the template ``\texttt{A sign that says `text'}'', scored by normalized edit distance:
    \begin{equation}
        r = \max\!\big(1 - N_e / N_{\mathrm{ref}},\, 0\big),
    \end{equation}
    where $N_e$ is the character-level edit distance between the recognized text and target string of length $N_{\mathrm{ref}}$.
    \item \emph{Human Preference Alignment}: PickScore~\cite{kirstain2023pick}, trained on large-scale side-by-side user preferences, measures visual appeal and prompt--image alignment. For scaling experiments on the 5.6B-parameter FLUX.3 Klein Base~\cite{flux3klein}, we employ HPSv3++~\cite{liu2026hpsv3pp}, a wide-spectrum preference reward model calibrated across model capability levels and RL iterations.
\end{itemize}

\spara{Implementation and Evaluation Protocol.}
Main benchmark comparisons are conducted at $512\times 512$ resolution on SD3.5-M~\cite{sd3}. Both the trainable policy velocity network $v_\theta$ and the online auxiliary velocity network $v_\phi$ are parameterized with lightweight LoRA adapters~\cite{hu2022lora}, while the pretrained base model serves as the frozen reference velocity $v_{\mathrm{ref}}$. During each iteration (\cref{alg:itdm}), we sample $G$ rollouts per prompt using deterministic ODE sampling, compute group-standardized rewards centered by the group mean baseline $b(\rvc)=\frac{1}{G}\sum_{i=1}^G R_i$, and apply advantage clipping $A_i = \operatorname{clip}(\widehat{\operatorname{Adv}}_i, -5, 5)/5$. 
By default, each optimizer update uses $24$ prompts with a group size of $G=24$; when a baseline exhibits training instability (e.g., AWM and NFT in \cref{fig:online_cfg1_hours}), we double the number of prompts per update ($48$ prompts) to stabilize training. 
Noisy states $\rvx_t$ are constructed by forward re-noising terminal clean latents via \cref{eq:path}. 
In the frozen-rollout offline regime (\cref{tab:drawbench_offline}), behavior rollouts are generated solely from the frozen pretrained SD3.5-M ($\text{CFG}=4.5$), with $G=2$ used for fair head-to-head comparison against Diffusion-DPO~\cite{wallace2024diffusion}. 
For evaluation on SD3.5-M, both task metrics and out-of-domain DrawBench~\cite{saharia2022photorealistic} metrics are computed using 40-step ODE sampling of the selected checkpoints under the corresponding guidance scale ($\text{CFG}=4.5$ or $\text{CFG}=1.0$). For the 5.6B FLUX.3 Klein Base~\cite{flux3klein} (\cref{tab:flux3-itdm-drawbench}), we perform online guidance-free post-training ($\text{CFG}=1.0$) with HPSv3++~\cite{liu2026hpsv3pp} and evaluate on DrawBench against the pretrained base model under both guidance-free ($\text{CFG}=1.0$) and default guided ($\text{CFG}=3.5$) inference.

\spara{Details of Out-of-Domain Evaluation Metrics.}
To detect reward hacking and verify general visual fidelity, we report seven unseen metrics on DrawBench:
1) \textbf{Aesthetic Score}~\cite{schuhmann2022laion}, a CLIP-based linear estimator trained on LAION aesthetic ratings;
2) \textbf{DeQA}~\cite{deqa}, a multimodal LLM assessor sensitive to low-level distortions, blurriness, and synthesis artifacts;
3) \textbf{CLIPScore}~\cite{hessel2021clipscore}, measuring semantic cosine similarity between prompt and image embeddings;
4) \textbf{HPSv2.1}~\cite{wu2023human}, a fine-tuned human preference model for stylistic and compositional alignment;
5) \textbf{ImageReward}~\cite{xu2023imagereward} (\textbf{ImgRwd}), evaluating text--image coherence, structural fidelity, and harmlessness;
6) \textbf{PickScore}~\cite{kirstain2023pick} on unseen DrawBench prompts (testing out-of-domain preference for GenEval/OCR models and unseen-prompt generalization for PickScore models); and
7) \textbf{UnifiedReward}~\cite{wang2025unified} (\textbf{UniRwd}), a unified multimodal human-preference reward model.

\end{document}